\documentclass[10pt,letterpaper]{article}

\usepackage{cogsci}
\usepackage{pslatex}
\usepackage[natbibapa]{apacite}
\usepackage{natbib}
\usepackage{amsmath}
\usepackage{caption}
\usepackage[dvipsnames]{xcolor}
\usepackage{graphicx}
\graphicspath{{./figs/}}
\usepackage{subcaption}
\usepackage{tabularx}
\usepackage{hhline}
\usepackage{multirow}
\usepackage{float}

\newcommand{\Table}[1]{Table~\ref{#1}}

\newcommand{\Figure}[1]{Figure~\ref{#1}}

\newcommand{\ignore}[1]{}

\newcommand\jp[1]{\textcolor{green}{#1}}

\newcommand\ie{\emph{i.e.}}
\newcommand\eg{\emph{e.g.}}

\newcommand\sub{Sub} 
\newcommand\basic{Basic} 
\newcommand\basicwpresub{Basic$_{sub}$} 
\newcommand\subwprebasic{Sub$_{basic}$} 

\title{Learning Hierarchical Visual Representations in Deep Neural Networks\\Using Hierarchical Linguistic Labels} 

\author{
\textbf{Joshua C. Peterson, Paul Soulos, Aida Nematzadeh, \& Thomas L. Griffiths} \\
Department of Psychology, University of California, Berkeley \\
\texttt{\{jpeterson,psoulos,nematzadeh,tom\_griffiths\}@berkeley.edu}}

\begin{document}

\maketitle

\begin{abstract}
Modern convolutional neural networks (CNNs) are able to achieve human-level object classification accuracy on specific tasks, and currently outperform competing models in explaining complex human visual representations. However, the categorization problem is posed differently for these networks than for humans: the accuracy of these networks is evaluated by their ability to identify single labels assigned to each image. These labels often cut arbitrarily across natural psychological taxonomies (e.g., dogs are separated into breeds, but never jointly categorized as ``dogs''), and bias the resulting representations. By contrast, it is common for children to hear both {\it dog} and {\it Dalmatian} to describe the same stimulus, helping to group perceptually disparate objects (e.g., breeds) into a common mental class. In this work, we train CNN classifiers with multiple labels for each image that correspond to different levels of abstraction, and use this framework to reproduce classic patterns that appear in human generalization behavior.
\end{abstract}


\section{Category Learning in Children and AI}
\label{sec:intro}



One of the challenges for a child learning a language is identifying what level in a hierarchical taxonomy a word -- category label -- refers to. After hearing the word ``dog'' upon observing a Dalmatian called Sebastian, a child needs to learn whether the category label ``dog'' refers to only Sebastian, all the Dalmatians, different breeds of dogs, etc. 
Psychologists have extensively studied this problem. A classic line of psychological research examined what level of abstraction (\eg, dogs vs. Dalmatians) conveys the most information; this level is called the \emph{basic level} \citep{rosch.1973,rosch.1976}. Another line of research has investigated whether people have an innate or a learned basic-level bias, \ie, a tendency to generalize a new category label to the members of the basic level of the taxonomy \citep[\eg,][]{markman.1991,golinkoff.etal.1994}.

Artificial intelligence (AI) systems need to address a similar problem. Given an image of a dog, an AI system needs to find the best label to describe that image. In computer vision, this problem is often formulated as an image classification task where the training data consists of images paired with single labels from a limited set of categories; a model needs to learn to predict the correct label for new images. Deep convolutional neural networks have been very successful at both achieving  state-of-art accuracy in the image classification task \citep{lecun2015deep} as well as learning representations that best explain human psychological and neural representations for natural images \citep[\eg,][]{mur2013human,agrawal2014pixels,JoshJosh&TomDeepSim2016}.

However, this problem formulation is different from what children experience. Multiple labels (\eg, Dalmatian and dog) are commonly used to refer to the same entity in the world children encounter. This explicit use of multiple category labels can help children learn a better representation of the taxonomic relations between categories -- the implicit hierarchical structure underlying the world. Different breeds of dogs form a category not only because of their perceptual similarity but also because they are referred to by the same label ``dog'', implicitly defining a higher level in the taxonomy. Moreover, forming a hierarchical representation can in turn help children better generalize to new items; for example, a new breed of dog, such as a poodle, will be categorized as a dog because of its similarity on crucial dimensions (e.g., long snout) to the members of that category. 

In this work, we investigate whether using such human-like supervision (labels from different levels of a hierarchy for a given entity) can help deep neural networks learn better visual representations. We explore the consequences of this multi-level classification by training a near state-of-the-art image classifier \citep{szegedy2015rethinking} on a dataset that provides multiple labels for images, each corresponding to a different level of the hierarchical taxonomy. In particular, we focus on two sets of labels: basic and subordinate (in the hierarchical taxonomy, subordinate labels, such as ``Dalmatian'', are categories that are below the basic level, such as ``dog'').
\ignore{
In this work, we explore the consequences of multi-level classification by training a near state-of-the-art image classifier \citep{szegedy2015rethinking} on a dataset that provides multiple labels for images, each corresponding to a different level of the hierarchical taxonomy. In particular, we focus on two sets of labels, basic and subordinate; subordinate labels (such as ``Dalmatian'') are categories that are below the basic level (such as ``dog'') in the hierarchical taxonomy.
} 

To examine the learned representations, we perform three sets of experiments. First, we explore whether the representations learned by each model capture the similarity relations observed in a hierarchical taxonomy. For example, we expect that different breeds of dogs will be better clustered together. We observe that training with basic-level labels results in both better grouping of similar examples and learning of hierarchical structure. 
Dendrograms reveal stark differences in hierarchical representations learned from basic as opposed to subordinate labels, resulting in interesting high-level groups not present in the original representations. Second, we show that the representations learned by training with basic-level labels match the performance of more commonly used subordinate-level label training schemes in explaining human similarity judgments, indicating that simpler classification tasks are sufficient to capture rich structure in human mental representations. 
Finally, we show that training with both label sets results in a better match to a classic effect of basic-level generalization bias observed in people using a simple model that restricts generalization with the number of consistent examples, an effect not previously replicated with a fully learned representation.

\ignore{
Take-away messages:
\begin{itemize}
\item The existing models exhibit basic-level generalization without explicit supervision but training on basic-level labels produce more human-like representations
\item Considering the right level of generalization for a given problem: train on basic-level as opposed to sub.
training on basic-level is cheaper
\end{itemize}
}

\section{Background}

\subsection{Exploring Levels of Generalization}
Cognitive scientists have extensively studied the problem of finding the correct level of generalization for a given label. The seminal work of \citet{rosch.1973} emphasizes the importance of basic-level objects because they are categorized first during perception and named early during word learning. 
Research on child word learning suggests that children have a bias in generalizing a new label to the the basic-level category members   \citep[\eg,][]{markman.1991,golinkoff.etal.1994}. 

\citet{xu2007word} further examined the basic-level bias by studying how children and adults generalize a new word after observing a few examples. They found that after observing examples from a subordinate level (\eg, three Dalmatians) labeled with a new word such as ``dax'', the participants often include dogs other than Dalmatians (\eg, poodles) in the category ``dax'' but exclude animals other than dogs (\eg, cats).
Previous models have successfully predicted the generalization patterns observed in this experiment; however, they require the use of pre-specified rather than learned hierarchical taxonomies based on adult similarity judgments \citep{xu2007word}, or limited feature representations \citep{nematzadeh.etal.2015.emnlp}. 
The main difference between our work and the existing models is that we explore whether the hierarchical structure of a taxonomy can be learned simply by jointly classifying natural images with multiple labels (or by choosing the right labels), mirroring the multiple references heard by children.

\subsection{Explaining Human Behavior with Deep Networks}
Another line of related work explores how representations from deep neural networks can be used to explain complex human behavior. For example, deep convolutional neural networks (CNNs; \citealp{lecun2015deep}) are specialized for image processing tasks, and can explain human typicality \citep{lake2015deep} and similarity ratings \citep{JoshJosh&TomDeepSim2016}, object memorability \citep{dubey2015makes}, and shape sensitivity \citep{kubilius2016deep}. \citet{JoshJosh&TomDeepSim2016} found that deep representations could be tuned post-hoc to improve fit to human similarity judgments by almost 50\%. Although the resulting representations better capture the human mental representations, the size of the image dataset used was relatively small, defining a limited context. Further, the reason for the initial discrepancies (and different stimulus groupings) is unknown, and the nature of the learned tuning is opaque. For this reason, it may be useful to instead consider training schemes that more closely match the pressures human categorizors face to more elegantly derive the relevant representations.

Recent work in explainable AI has emphasized the importance of understanding the biases that the training data or a model's architecture can introduce \citep[\eg,][]{zhao2017men, ritter2017cognitive}. Our analysis sheds light on another one of these important biases -- the basic-level bias -- that the image classification model captures.

\ignore{
\citet{wang2015basic} were the first to take inspiration from \citet{rosch.1973} to improve CNN classifiers. train CNN classifiers on a set of $308$ basic level labels collected for the $1000$ ILSVRC12 classes. They use this network as an initialization that is then fine-tuned on the original 1000 classes, obtaining higher accuracy than their baseline AlexNet model trained without basic level pre-training.
Our work is different from these in that we assess the mutual pressure between potentially competing basic and subordinate classification objectives, inspect the representations, and probe the implications on generalization after training.
}

\subsection{Multi-level and Multi-label Classification}
Computer vision research has recognized the importance of training multi-label classifiers to take advantage of the available rich linguistic information in tasks such as image understanding \citep[\eg,][]{wang2016cnn}. Some of this work makes use of labels from different levels of a hierarchy, but has mostly focused on improving classification performance; for example, \citet{lei2017weakly} show that using coarse-grained labels can improve the classification accuracy of finer-grained categories. 
\citet{wang2015basic} are the first to take inspiration from the work of \citet{rosch.1973}, but they also focus on improving subordinate-level benchmarks. They trained CNN classifiers on a set of $308$ basic level labels that encompassed an existing set of $1000$ ILSVRC12 competition classes \citep{russakovsky2015imagenet} that clearly fit the traditional conception of subordinate categories \citep{rosch.1976}. They used this network as an initialization that was then fine-tuned on the original 1000 classes, obtaining higher accuracy than their baseline model trained without basic level pre-training.
Our work differs in that we assess the structure of the learned representations and implications on later generalization behavior, both for each label set as the sole source of supervision, and as simultaneous and potentially competing objectives.

\begin{figure*}[!ht]
  \begin{center}
  \includegraphics[width=1.0\linewidth]{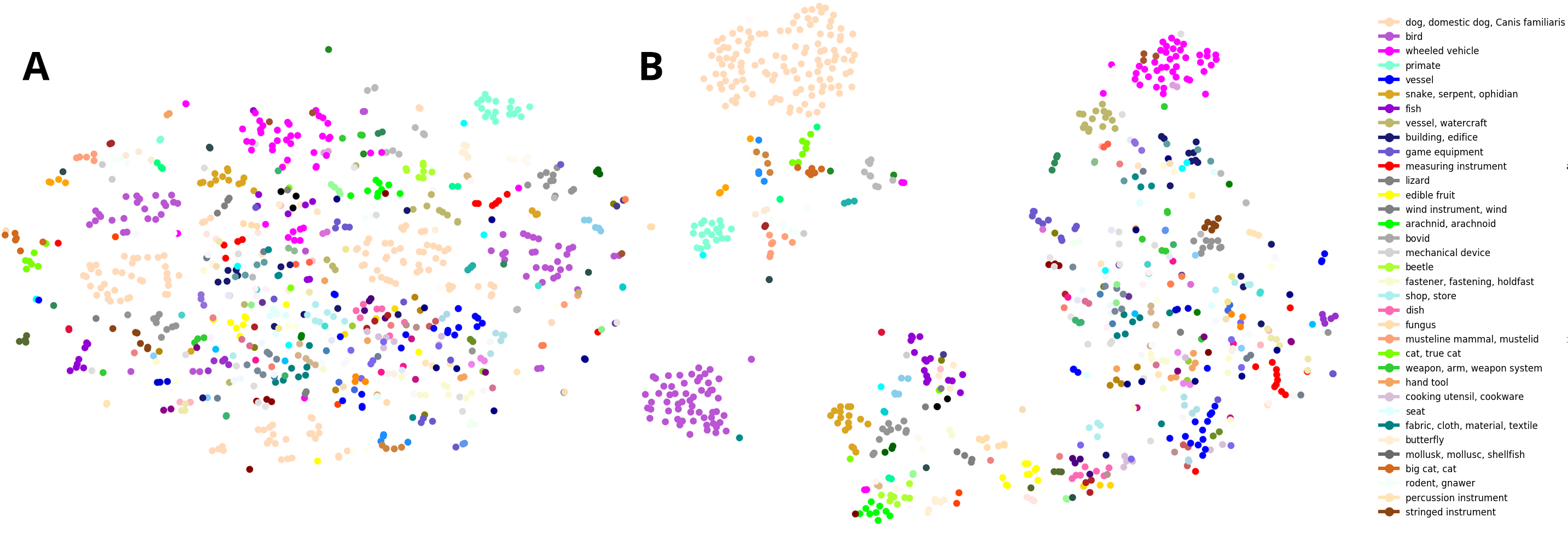}
  \caption{t-SNE visualizations of representations learned for (\textbf{A}) \sub\ training and (\textbf{B}) \basic\ training, colored by each of the 35 basic labels with the most subordinate classes for clarity. The basic model has tighter clusters and more distinct boundaries between categories.}
  \label{fig:tsne_plots}
  \end{center}
  \vspace{-5mm}
\end{figure*}

\section{Training Paradigms}

We are interested in examining the extent to which linguistic labels from different levels of the taxonomy (\ie, basic and subordinate categories) affect the structure and hierarchical taxonomy of the learned representations. To examine this hypothesis, we train classifiers using each level separately, and also using multiple labels (first taking a pretrained model on one level and tuning it using labels from both levels). Our goal in the latter case is to more closely mimic a child's experience in language learning -- children are known to make use of simple labels first (\eg, ``dog''; \citealp{rosch.1976}), later followed by increasingly sharper distinctions (\eg, ``poodle'').\footnote{Another possibility is to train multi-label classifier simultaneously on the two levels which we will explore in future.}

More specifically, we pose the multi-level labeling problem simply as learning a set of independent softmax classifiers that are unconnected to each other and fully connected to the final representation layer of a deep CNN. We define the loss function for the multi-level classifiers as $0.5\mathcal{L}_{basic}+0.5\mathcal{L}_{subordinate}$. Because these models have already been pretrained on one level, this joint loss can be thought of as a method to prevent the fine-tuning procedure from overwritting previous knowledge about the original training domain (\ie, we would like the model to remember what it learned about basic labels while approaching the new problem of more fine-grained subordinate labeling, and vice versa).
While other alternative approaches exist for defining the network architecture and loss function, this approach provides a single embedding space for all images, which allows us to inspect the representations with classic psychological methods such as hierarchical clustering.

We compare the representations from our multi-level classifiers with ones that are trained using only one level (basic or subordinate). To train on single levels, we use a $k$-way softmax classifier as the final layer, where $k$ is the number of classes for that level. We report the results for four different models:

\begin{itemize}
    \vspace{-0.2cm}
    \item \textbf{\sub}: Trained only on subordinate.
    \vspace{-0.3cm}
    \item \textbf{\basic}: Trained only on basic.
    \vspace{-0.3cm}
    \item \textbf{\subwprebasic}: Pretrained on basic; tuned with subordinate.
    \vspace{-0.3cm}
    \item \textbf{\basicwpresub}: Pretrained on subordinate; tuned with basic.
    \vspace{-0.2cm}
\end{itemize}

To train our multi-level classifiers, we use the model architecture defined in InceptionV3 \citep{szegedy2015rethinking} since it obtains high accuracy while still being relatively quick to train. It contains $159$ layers, $23$ million parameters, and a top-1 validation accuracy of nearly 79\% on \textsc{ILSVRC12} using the default subordinate label set. We use the off-the-shelf pretrained version of this network for the \sub\ model.
For the \basic\ model, 
we reinitialize the network parameters randomly and retrain from scratch. For fine-tuning models, we freeze all but the weights in the last block of InceptionV3 to speed training.
%
%

Following \citet{wang2015basic}, we use the $1000$ labels from \textsc{ILSVRC12} as our subordinate classes, and basic level labels provided by the same authors, described in the previous section.

\section{Exploring Representations}

We perform a set of exploratory qualitative visual analyses to examine whether the representations learned by each model reflect the grouping of similar objects observed in a hierarchical taxonomy. If representations capture such higher-level abstractions, not only would examples of Dalmatians cluster close to each other, but also different breeds. To analyze the learned representations, we extracted a 2048-dimensional feature vector from the last hidden layer before the output, and used for all subsequent analyses in this paper.

We first explored the representations using t-SNE \citep{maaten2008visualizing}, a common visualization method for CNN representations, shown in Figure \ref{fig:tsne_plots}. 
The t-SNE method reduces the 2048-dimensional vectors to 2 dimensions so that we can visually compare learned representations. We find that the \sub\ model does not appear to cluster basic level categories very well (\eg, there are multiple spatially separated clusters of dogs), whereas the \basic\ model does a much more effective job of clustering these like subordinate classes together (\eg, dog breeds are now unified under a single tight cluster). Models trained on both label sets simultaneously, \subwprebasic\ and \basicwpresub) exhibited similar clustering to the \basic\ model, and so they are omitted for this reason.

\begin{figure*}
  \begin{center}
  \includegraphics[width=1.0\linewidth]{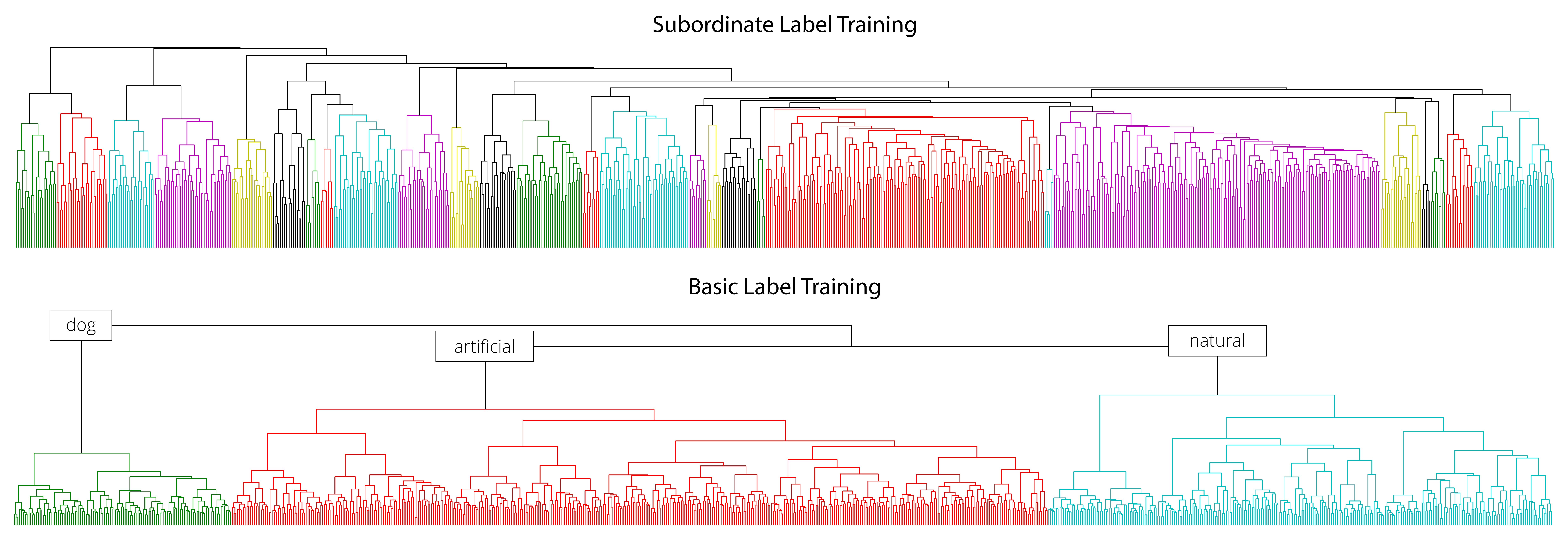}
  \caption{Dendrograms showing the learned representations. The model trained on subordinate labels (top) has no clear hierarchical structure. The model trained on basic labels (bottom) has a clearly defined hierarchical structure which divides the validation data into three top level categories: dogs, images of artificial items, and images of natural items. Over 10\% of Imagenet is images of dogs which is why there is a separate branch specifically for these images. We did not label the subordinate model because it did not have a definable structure.}
  \label{fig:dends}
  \end{center}
  \vspace{-5mm}
\end{figure*}

We also plot dendrograms of the the representations in Figure \ref{fig:dends}. The \basic\ model has a clearly defined hierarchy, whereas the \sub\ model does not. The branch we call ``artificial'' contains images showing artificial objects such as cars, buildings, household objects, sports, and technology. The other main branch, the ``natural'' distinction show animals, fish, mushrooms, and other natural stimuli. Interestingly, this high-level distinction is not present explicitly in the basic label set, and is one of the defining categorical divisions found in human mental representations \citep{mur2013human}. As before, models trained on both label sets (\subwprebasic\ and \basicwpresub) behaved similarly to the \basic\ model, suggesting that significant basic-level supervision at any stage of training allows for the preservation of basic-level clustering.

\section{Predicting Human Similarity Judgments}

Another way to evaluate the goodness of the learned representations is examine how well they predict human psychological representations as captured by pairwise human similarity judgments \citep{shepard1987toward}.
As a gold-standard dataset, we use the similarity ratings collected by \citet{JoshJosh&TomDeepSim2016}. This dataset consists of $7,140$ pairwise comparisons of all $120$ images presented to MTurk workers, who were asked to rate their similarity between 0 and 10.

For each model, we compute the similarity between image pairs as the inner product of their learned representations. Then, we calculate the Pearson correlation between the representations learned by each model and the gold-standard similarity ratings. We report $R^2$ to compare directly with previous work (see \Table{tb:r_sqr}).
Interestingly, the performance of the models that either train solely on basic labels or are tuned with basic labels greatly surpass those that train solely on subordinate labels or are tuned with subordinate labels. 
%
%
We note that previous work has obtained very similar correlations to our \basic\ results by using a different (yet comparable) network trained only on subordinate labels \citep{JoshJosh&TomDeepSim2016}. This may indicate the current network is generally less apt to predict human similarity judgments, but is more likely to encode information relevant to people given the right supervision from basic labels. 
In addition, our results show that human psychological representations can be approximated using simpler classifiers with more coarse-grained object distinctions (basic level labels).
It is also worth noting that most of the previous work in explaining human visual representations uses subordinate-label trained networks \citep{agrawal2014pixels,mur2013human}; our results show a new alternative for predicting these representations.

\begin{table}
\begin{center}
\begin{tabular}{ccccc}
\hline
& \sub & \basic  & \subwprebasic & \basicwpresub \\
\hline
    $R^2$ & 0.38 & 0.57 & 0.41 & 0.57 \\
\end{tabular}
\caption{Variance explained in human similarity judgments by representations formed by each model.}
\label{tb:r_sqr}
\end{center}
\vspace{-5mm}
\end{table}

\ignore{ 
\begin{center}
\begin{table}
\begin{tabular}{ c |c }
Basic Label Training & Subordinate Label Training \\
\hline
     $0.57$ & $0.38$   \\ 
\end{tabular}
\caption{The Pearson correlation between similarity judgments of each model and humans.}
\label{tb:pearson}
\end{table}
\end{center}

}
\section{Generalization Experiments}

\begin{figure}[!ht]
   \begin{center}
   \includegraphics[width=\linewidth]{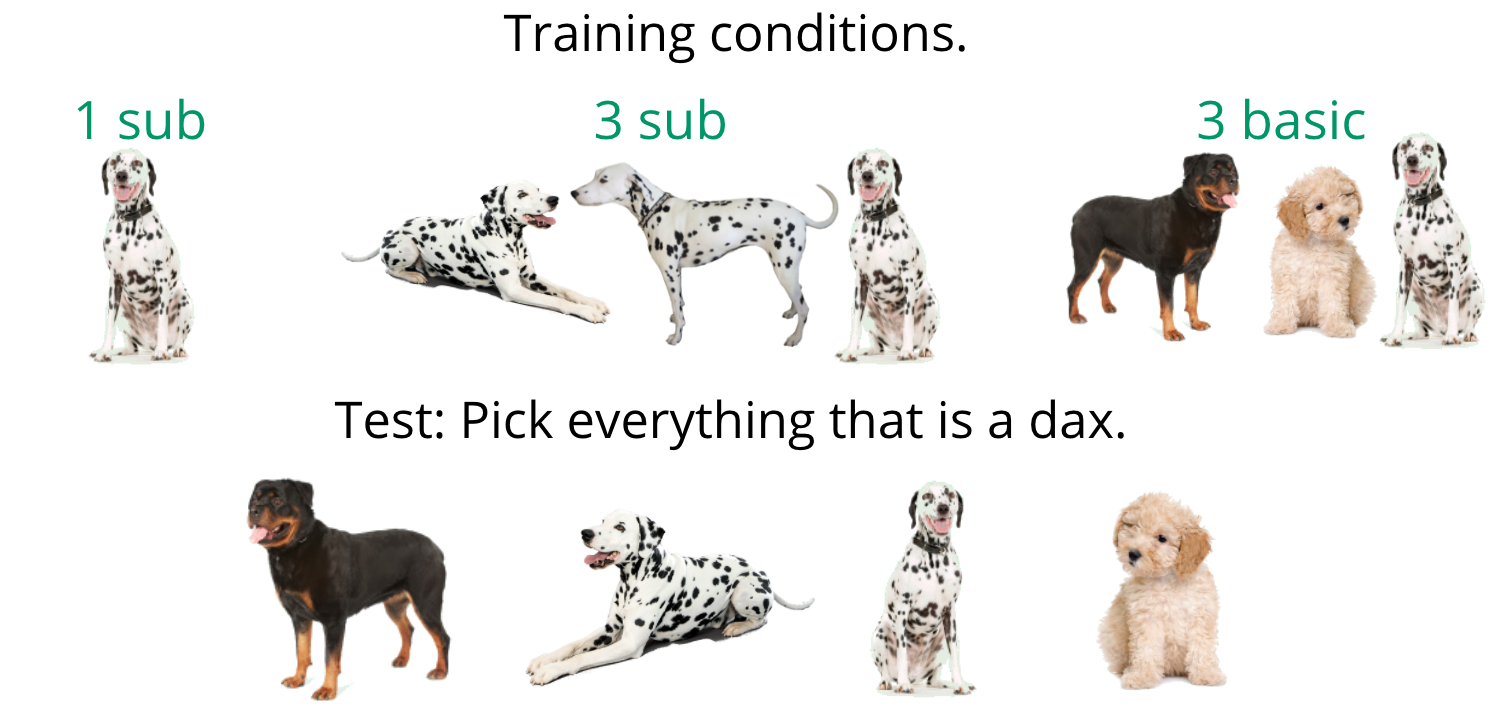}
   \caption{The setup of X\&T's experiments \cite{xu2007word} used in this paper. The first row shows the three training conditions and the second row is an example of the test trial. Note that during the test trial, the basic matches (dog breeds other than Dalmatians) are different from the ones observed during training (in contrary to what is shown here).}
   \label{fig:xtsetup}
   \end{center}
   \vspace{-5mm}
\end{figure}

\citet{xu2007word} (henceforth, X\&T) examined how people generalize a novel word label after observing a few examples, and whether the number or the taxonomic level from which the examples were drawn changes the generalization behavior. For example, the participants heard a word label such as ``dax'' while observing one Dalmatian, three Dalmatians, or three different breeds of dogs. \Figure{fig:xtsetup} shows the set-up of their experiments.

The experiment set-up of X\&T's can be thought of as a few-shot learning task.
We examine whether our models exhibit the generalization behavior observed in these experiments. We focus on three of their training conditions. 
\vspace{-1mm}
\begin{itemize}
\item \textbf{1 sub}: the model (or a participant in X\&T's experiment) observes 1 (subordinate) example, such as a Dalmatian.  
\vspace{-0.3cm}
\item \textbf{3 sub}: the model receives 3 examples from the same subordinate category, such as 3 Dalmatians.
\vspace{-0.3cm}
\item \textbf{3 basic}: 3 examples are drawn from a basic-level, such as a poodle, a Dalmatian, and a Great Pyrenees. 
\end{itemize}
\vspace{-1mm}
In X\&T's experiments, after training, the participants were asked to pick everything that is a ``dax'' from a fixed set of examples drawn from different levels of taxonomy. We focus on the test objects that are relevant to our experimental conditions, \ie, two basic-level and two subordinate matches.\footnote{X\&T also included superordinate examples (such as an animal other than a dog). Here we only focus on the subordinate and basic-level matches because our training data only includes those labels.} For example, after observing one Dalmatian as a ``dax'', the participants had to decide whether two other Dalmatians, a poodle, and a golden retriever are also examples of ``dax''. The results of their experiment is shown in \Figure{fig:generalization}. Two interesting observations can be made from their results: First, people exhibit basic-level bias since they generalize to the basic-level after observing only subordinate examples. Second, their degree of basic-level generalization decreases after observing 3 examples (see ``Basic Match'' in 1-sub and 3-sub conditions in plot titled ``Human'' in \Figure{fig:generalization}).

We simulate the different conditions of X\&T's experiments as a few-shot generalization task, and examine the generalization behavior of models on both subordinate and basic-level matches.

\subsection{Learning from Positive Examples}

To mimic generalization experiments with humans, we use a k-shot generalization formulation meant to learn concepts from positive examples only. Specifically, we use exponentiated euclidean distance (in the spirit of \cite{shepard1987toward}, given that we are interesting in making qualitative comparisons to human behavior), normalized over distances to all items in the test set to obtain generalization probabilities:

\vspace{-2mm}
\begin{equation}
g(q_i,c) = \frac{e^{-d(q_i,c)}}{\sum_j e^{-d(q_j,c)}},
\vspace{-1mm}
\end{equation}
where $g$ is the generalization function, $c$ is concept template (a single training example or the mean of several training examples), $q_i$ is the query (test) image, and $d$ is the euclidean distance function described above.

It has been demonstrated that human generalization behavior tightens as the number of examples increases \citep{tenenbaum2000rules}. We can represent this phenomenon in our model through a simple augmentation:

\vspace{-2mm}
\begin{equation}
g'(q_i,c) = \frac{e^{-nd(q_i,c)}}{\sum_j e^{-nd(q_j,c)}},
\vspace{-1mm}
\end{equation}
where $n$ is the number of positive examples observed. 

Since we care only about the relative differences in generalization probabilities across different levels of testing stimuli, we further normalize each set of probabilities for each training set by dividing each probability by the largest $g'(q_i,c)$ in the set (e.g., the largest probability becomes 1).

To evaluate this model, we sample test and train stimuli analogous to X\&T from $88/308$ of the basic level labels in our set that encompassed at least $3$ subordinate classes, a requirement for the final three-example experiment condition.

\subsection{Generalization Results}
\begin{figure*}[!ht]
  \begin{center}
  \includegraphics[width=1.0\linewidth]{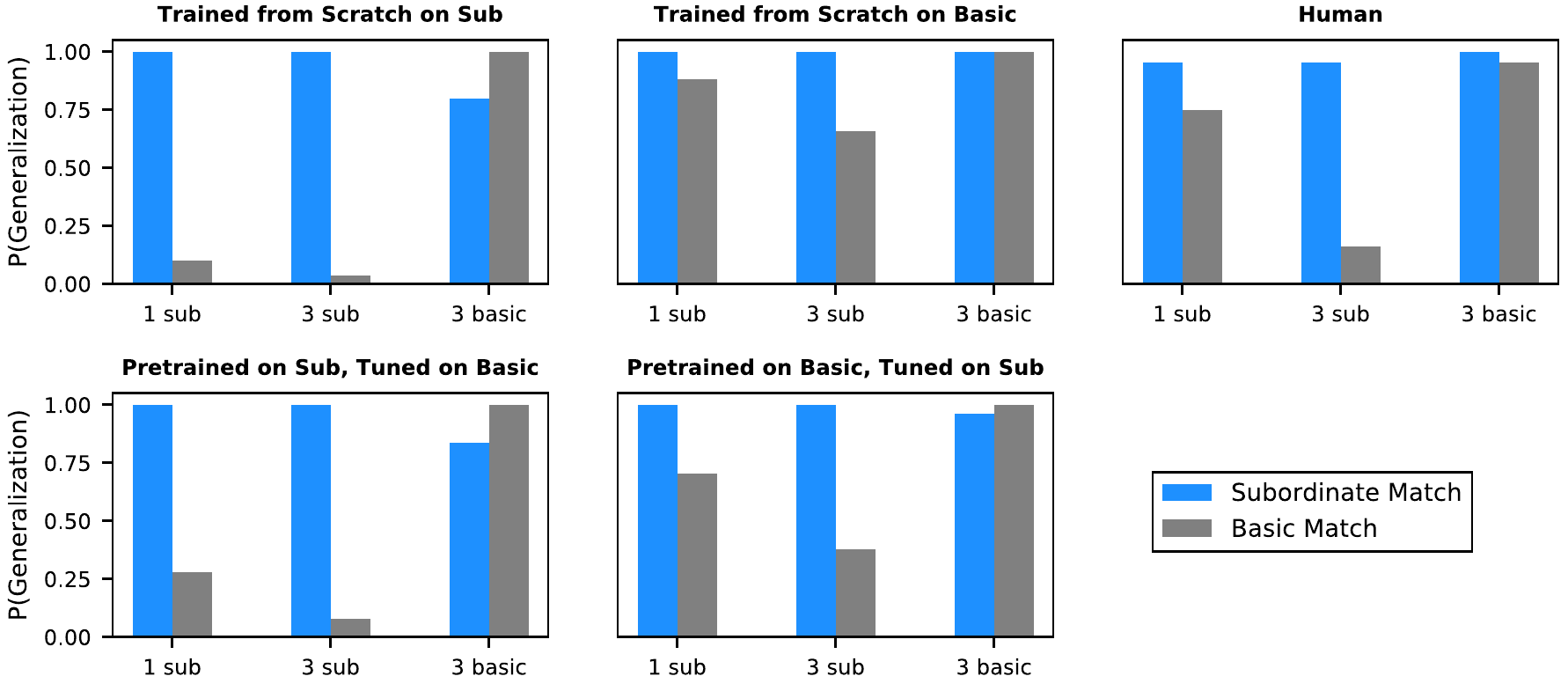}
  \vspace{-6mm}
  \caption{The results of generalization experiments from our trained models and the human data from X\&T's experiments.}
  \label{fig:generalization}
  \end{center}
  \vspace{-6mm}
\end{figure*}

The results of the generalization experiments from all four of our models are shown in \Figure{fig:generalization} (see top left and all bottom plots).
To some extent, all models exhibit a basic level bias (there is some non-zero probability assigned to generalization at the basic level, even when a single example is given). However, this effect is extremely weak in the \sub\ model. In all other models that involve basic labels, a stronger basic-level bias is apparent after training on 1 example or 3 subordinate examples, strongly resembling more human-like trends. In addition, this basic-level generalization decreases as the model processes more examples (compare 1 sub to 3 sub conditions).
Moreover, most models also exhibit subordinate and basic-level generalization after observing 3 examples drawn from a basic-level category, further resembling humans.
Note that the observed basic-level bias for the \sub\ training, just by capturing the perceptual similarities present in the data, to some extent learns about the higher-level similarities (groupings) of the examples. However, adding nearly any form of higher-level supervision (\ie, using basic-level labels) produces higher basic-level generalization and a better match to human behavior. We also observe a basic-level bias with $g$ (not shown), but without incorporating the number of examples into the formulation, the basic-level generalization is larger than that of people in the 3-sub condition.

\section{Conclusion}

Image classification models are often trained on images paired with single labels that correspond to the subordinate level of the hierarchical taxonomy. People, on the other hand, use multiple labels to refer to the same entity in the world (\eg, dog and Dalmatian). This explicit use of multiple labels makes the category learning problem easier as people can use the labels to identify similarities between the concepts (in addition to other features, such as perceptual cues, that signal similarity).
Moreover, the use multiple labels helps form hierarchical representations that capture different levels of abstraction. For example, two Dalmatians are more similar to each other than a Dalmatian to a poodle; dalmatians and poodles are still highly similar and should share more relevant features than a Dalmatian and a cow (body size as opposed to black and white spots).
This hierarchical nature of the representation is in turn helpful in categorizing and labeling a new entity: a child can use the similarity of an unobserved breed of dog (a Great Pyrenees) to previously-seen dogs to categorize and infer a label for it.

In this paper, we show that training on basic-level and sub-ordinate labels (or just basic-level labels) results in representations that better capture the hierarchical structure of taxonomy of real-world objects.
We also show that these representations result in a better match to the basic-level bias observed in human generalization behavior.

Interestingly, we observe that a relatively big portion of ImageNet ($10\%$) are subordinate dog labels. Future work should look at other datasets with more interesting, less skewed basic label stratifications. Additionally, we only explore basic and subordinate levels in this work, which are both known to share many features in common with their members \citep{rosch.1978}, whereas high-level superordinate classes (e.g., furniture) often have members with few shared features.

\smallskip
\noindent \textbf{Acknowledgements.} This work was supported by grant number 1718550 through the National Science Foundation.


\renewcommand{\bibliographytypesize}{\small}
\bibliographystyle{apacite}
\setlength{\bibleftmargin}{.125in}
\setlength{\bibindent}{-\bibleftmargin}
\bibliography{citations}

\end{document}